\pdfoutput=1

\documentclass[11pt]{article}

\usepackage[preprint]{coling}

\usepackage{times}
\usepackage{latexsym}

\usepackage[T1]{fontenc}

\usepackage[utf8]{inputenc}

\usepackage{microtype}

\usepackage{inconsolata}

\usepackage{graphicx}

\usepackage{amsmath}
\usepackage{siunitx}
\usepackage{booktabs}
\usepackage{multirow}

\usepackage{xcolor}
\usepackage{soul}

\title{Linguistic Features Extracted by GPT-4 Improve Alzheimer's Disease Detection based on Spontaneous Speech}

\author{Jonathan Heitz \\
   Department of Psychology, University of Zurich \\Methods of Plasticity Research, Zurich, Switzerland \\
  Language & Medicine Competence Centre, University of Zurich, Zurich, Switzerland \\
  \texttt{email@domain} \\\And
  Gerold Schneider \\
  Department of Computational Linguistics \\ University of Zurich, Zurich, Switzerland \\
  \texttt{email@domain} \\\And
  Nicolas Langer \\
  Department of Psychology, University of Zurich \\ Methods of Plasticity Research, Zurich, Switzerland \\
  \texttt{email@domain} \\}

 \author{Jonathan Heitz\textsuperscript{1,3}, Gerold Schneider\textsuperscript{2,3}, Nicolas Langer\textsuperscript{1} \\
         \textsuperscript{1} Department of Psychology, University of Zurich, Methods of Plasticity Research, Zurich, Switzerland\\
\textsuperscript{2}  Department of Computational Linguistics, University of Zurich, Zurich, Switzerland\\
\textsuperscript{3}  Language \& Medicine Competence Centre, University of Zurich, Zurich, Switzerland \\
         jonathan.heitz@uzh.ch, gschneid@cl.uzh.ch, n.langer@psychologie.uzh.ch}

\begin{document}
\maketitle
\begin{abstract}
    Alzheimer's Disease (AD) is a significant and growing public health concern. Investigating alterations in speech and language patterns offers a promising path towards cost-effective and non-invasive early detection of AD on a large scale. Large language models (LLMs), such as GPT, have enabled powerful new possibilities for semantic text analysis.  In this study, we leverage GPT-4 to extract five semantic features from transcripts of spontaneous patient speech. The features capture known symptoms of AD, but they are difficult to quantify effectively using traditional methods of computational linguistics. 
    We demonstrate the clinical significance of these features and further validate one of them (``Word-Finding Difficulties'') against a proxy measure and human raters.
    When combined with established linguistic features and a Random Forest classifier, the GPT-derived features significantly improve the detection of AD. Our approach proves effective for both manually transcribed and automatically generated transcripts, representing a novel and impactful use of recent advancements in LLMs for AD speech analysis.
\end{abstract}

\section{Introduction}

In light of the global demographic shift towards an older population, Alzheimer's Disease (AD) emerges as a critical public health concern with a substantial economic burden \cite{weller2018current}. While there is no known cure, effective management depends on early diagnosis \cite{arvanitakis2019diagnosis}, necessitating the identification of biomarkers that are easy to collect, cost-effective, and non-invasive \cite{ribaldi2019large}. Speech and language alterations manifest as early symptoms of AD \cite{calza2021linguistic}, presenting a promising avenue for real-time AD screening through speech analysis in extensive epidemiological research. 
For these methods to be adopted on a large scale, it is imperative to develop fully automated, reliable, and explainable systems capable of providing real-time predictions.

The recent rise of large language models (LLMs) has opened new possibilities of automatic natural language processing. One of the most popular and powerful LLMs is OpenAI's GPT series, the most capable model being GPT-4, which excels at a variety of traditional language processing benchmarks and beyond, including e.g.\ the Medical Knowledge Self-Assessment Program \cite{openai2023gpt4}. 
Despite the widespread integration of GPT into various applications, the field of AD speech analysis remains relatively underdeveloped in comparison, with prevailing methodologies still predominantly reliant on conventional (acoustic and linguistic) features or older language models such as BERT \cite{bert2018} (c.f.\ \citet{parsapoor2023aibased} for a recent review on the topic). While these conventional approaches have shown impressive results distinguishing AD from control, they are ineffective in detecting some of the clinically known, but more complex and high-level symptoms of AD in speech, such as word-finding difficulties.

In this study, we address this gap by leveraging GPT in two distinct approaches for AD detection: a) fine-tuning a GPT model for direct classification purposes, and b) employing GPT as a rich semantic feature extractor from transcripts. The extracted features quantify complex speech alterations in AD, which existing methodologies are unable to capture. When combined with established features and Random Forest, they yield substantial improvements in AD detection, while enhancing explainability.

\section{Related Work}

Significant work has been done on AD classification based on spontaneous speech. Recently, the most popular datasets in the field were introduced as part of the ADReSS and ADReSSo challenges \cite{Luz2020ADReSS, luz21_adresso}. While some approaches leverage information encoded in the audio signal directly, most work has found the linguistic analysis of transcripts more useful \cite{cummins2020comparison}. Methodologically, most such studies can be divided into a more traditional approach of feature extraction paired with the use of popular classification algorithms (e.g.\ \citealp{tang2023explainable, taghibeyglou2024context}), or fine-tuned language models, most prominently BERT (e.g.\ \citealp{balagopalan2020bert, pan2021using}). The feature-based approach has the advantage of being more transparent and explainable. Fine-tuned language models, on the other hand, have been reported to produce slightly better classification performance, although the reported performance differences are usually small \cite{balagopalan2021comparing} and depend on the setup (e.g.\ manual vs.\ automatically generated transcripts, \citet{heitz2024influence}).

LLMs, in particular (Chat)GPT, have been applied in a variety of medical settings, including medical competency examinations \cite{nori2023capabilities} and diagnosis \cite{hirosawa2023diagnostic, wang2023can}. In the context of cognitive decline or AD, 
prior work has used GPT for data augmentation of speech transcripts \cite{cai2023exploring} or employed it to directly classify AD from control relying on a prompt listing participant demographic information and cognitive test scores  \cite{wang2023can}. Limited work has studied GPT on spontaneous speech transcripts: \citet{yang2023gpt, wang2023text} attempted to distinguish patients with mild cognitive impairment (MCI) from healthy controls, iteratively improving ChatGPT prompts. However, they used a non-standard and unbalanced dataset, did not compare performance to traditional approaches, and attempted direct (zero-shot) classification, making it impossible to combine GPT-extracted information with established features. \citet{balamurali2024performance} experimented with the ADReSSo dataset, but the results are only slightly better than the random baseline. Both of these approaches queried GPT via the web-based ChatGPT interface, which uses random sampling of responses, limiting reproducibility. 

The novelty of our contribution is three-fold: Firstly, to the best of our knowledge, this is the first study leveraging GPT to extract linguistic features from transcripts of spontaneous speech and integrating them into an existing pipeline. 
Secondly, we validate the GPT-extracted features: We assess their clinical significance through a group comparison, and further scrutinize one of them by measuring its alignment with a proxy measure and human evaluations.
Thirdly, our evaluation is more rigorous than prior work by using a standard dataset, querying OpenAI's GPT API with reproducible settings, assessing output stability to prompt and random seed variations, and comparing our approach to an established combination of linguistic features and Random Forest, as well as a fine-tuned GPT model. Our innovative use of GPT improves AD classification performance, with advantages in explainability.

\section{Methods}

All training and analysis is performed on a Linux Ubuntu machine with 8 CPUs, \SI{32}{\giga\byte} RAM, and a NVIDIA Tesla T4 GPU. Our pipeline is implemented in Python 3.12, and our code for preprocessing, feature extraction, model training, and evaluation is available on our GitHub repository\footnote{\href{https://github.com/jheitz/coling2025_gpt_paper}{https://github.com/jheitz/coling2025\_gpt\_paper}}. 

\subsection{Dataset and preprocessing}
In this work, we use the English ADReSS dataset \cite{Luz2020ADReSS}, containing audio recordings of 156 participants describing the Cookie Theft picture \cite{goodglass2001bdae}. The dataset is balanced with respect to diagnosis, age, and gender, and includes manual transcriptions in the CHAT annotation format \cite{macwhinney2000childes}. Basic demographic characteristics are provided in Table~\ref{tab:dataset}.

\begin{table}[bht]
    \centering
\scalebox{1.0}{
\begin{tabular}{llll}
\toprule
 & \multicolumn{1}{p{1.6cm}}{n (f, m)} & Age & MMSE \\
\midrule
AD & 78 (43, 35) & $66.6 \pm 7$ & $17.8 \pm 5.5$ \\
Control & 78 (43, 35) & $66.3 \pm 7$ & $29.0 \pm 1.2$\\
\bottomrule
\end{tabular}
}
\caption{Characteristics of the ADReSS dataset for AD patients and control subjects: We report the total number of subjects (n) as well as the number of female (f) and male (m) participants. In addition, mean and standard deviation are given for age and Mini Mental State Examination scores (MMSE, \citealp{folstein1975MMSE}).}
\label{tab:dataset}
\end{table}

This dataset is a subset of the DementiaBank English PITT corpus \cite{becker1994pittcorpus}, with noise removal applied to the audio files \cite{Luz2020ADReSS}. We find this noise removal problematic, as it also removes significant sections of speech, resulting in multiple audio files with no or very little intelligible participant voice. This renders downstream automatic speech recognition difficult. To counteract this problem, we matched the ADReSS selection of recordings with the original PITT audio files and use the latter in our approach. 

Some audio files contain interviewer speech sections, such as ``\emph{Is there anything else?}". These could bias the AD classification task, as they appear more frequently in AD patients than controls. To avoid any such interference, we remove interviewer utterances from the audio (using timestamps provided in the CHAT transcription file) and from the manual transcripts, leaving only the participant's speech. 

The manual transcripts provided by the dataset contain special annotations and transcription codes that go beyond the pure transcription, explicitly marking pauses, retractions, and fragments, among other things. We remove these extra annotations, while retaining all uttered words (including disfluencies such as ``uhm''). The result of this preprocessing is a pure word-by-word transcription, similar to one that might be produced by an Automatic Speech Recognition algorithm\footnote{Preprocessing is identical to our previous work and described there in detail (\citealp{heitz2024influence}, Appendix~A).}.

\subsection{Automatic Speech Recognition (ASR)}
To evaluate the effectiveness of our approach in a fully automatic AD detection pipeline, we compare the use of manual transcripts to results from automatic speech recognition (ASR). We employ two pre-trained state-of-the-art ASR models: Whisper \cite{radford2023robust} and the Google Speech ``Chirp'' model \cite{zhang2023googleums}, both of which have reported excellent performance\footnote{Whisper is run locally using HuggingFace's model \texttt{openai/whisper-large-v3} in the \texttt{transformers} library v4.37, while Google Speech is accessed via the Google Cloud API \texttt{Cloud Speech-to-Text V2}.}.

We assess the quality of these transcriptions using the Word-Error-Rate (WER) \cite{morris2004wer}, which quantifies differences between ASR transcripts and the manual transcripts provided as part of the dataset, and is defined as 
\begin{equation}
    WER = \frac{I + D + S}{N}
\end{equation}
where $N$ denotes the number of words in the manual transcription, and $I$, $D$, and $S$ count insertions, deletions, and substitutions of words.

\subsection{Feature engineering}
\subsubsection{Established features (baseline)}
As a baseline, we use a set of $40$ linguistic features which we call \emph{Established features}. These features include lexical features, features based on part-of-speech (POS) tagging, and features of repetitiveness, and they have worked well in previous approaches of AD classification on speech. The inclusion criteria, the list of features, and their definitions are detailed in Appendix~\ref{appendix:features}.

\subsubsection{GPT features}
\label{sec:gpt-prompts}
We prompt GPT-4 \cite{openai2023gpt4} programmatically via the OpenAI API to extract relevant features from a transcript. Reproducibility of GPT outputs is maximized by setting a seed and specifying ${\footnotesize\texttt{temperature}}=0$ in the API calls. \\

\noindent\textbf{Choice of features} (\emph{Prompt 1}): To initially identify relevant features, we query GPT once for five indicators of AD in spontaneous speech, without providing any transcripts. The GPT prompt and its response are provided in Appendix~\ref{appendix:gpt-prompts}. We call the resulting features \emph{GPT features}, and they consist of: \emph{Word-Finding Difficulties (Anomia)}, \emph{Semantic Paraphasias}, \emph{Syntactic Simplification}, \emph{Impoverished Vocabulary}, and \emph{Discourse Impairment}. The free-text descriptions of these features are also provided by GPT (they are contained in \emph{Prompt 2}, displayed in Figure~\ref{fig:gpt-prompt}). 

Note that all five indicators have previously been reported to occur more frequently in AD patients than controls \cite{forbes2005age,croisile1996comparative,forbes2002distinct,lindsay2021language}; in other words, GPT successfully identifies known symptoms of this disease, motivating its use in our application. \\

\noindent\textbf{Feature extraction} (\emph{Prompt 2}): For each transcript, we query GPT-4 to quantify each feature on a Likert scale from 1 (not at all fulfilled) to 7 (very strongly fulfilled), without providing information on the diagnostic label (AD or control). The scale from 1 -- 7 is motivated by the ``Rating Scale Profile of Speech Characteristics'' of the Boston Diagnostic Aphasia Examination (BDAE, \citealp{goodglass2001bdae}), where a similar scale is used by clinicians to rate characteristics of spontaneous speech in the context of aphasia. 
In addition to a numerical assessment, we ask GPT for text extracts from the transcript to explain the assessment. The prompt and an example output are given in Figure~\ref{fig:gpt-prompt}. 

\begin{figure}
    \centering
    \scalebox{0.7}{
\begin{minipage}{10.5cm}
\textbf{Prompt\,2:} \texttt{Here are key indicators of Alzheimer's dementia in spontaneous speech:} \\\\
\texttt{Word-Finding Difficulties (Anomia): Individuals with Alzheimer’s often have trouble finding the right words. This can manifest as frequent pauses, use of non-specific words like ``thing'' or ``stuff'', or circumlocutions (talking around the word without being able to name it). For example, instead of saying "stool”, a person with dementia might say ``the thing you sit on''.\\
Impoverished Vocabulary: The range of words used by someone with Alzheimer’s may become limited. Their language may seem less rich and more repetitive, with a reliance on common and general terms rather than specific nouns or verbs.\\
Syntactic Simplification: The complexity of sentence structures may decline. People with dementia might use simpler, shorter sentences and may make more grammatical errors. They might avoid complex grammatical constructions like subordination or relative clauses.\\
Semantic Paraphasias: This refers to the use of incorrect words that are semantically related to the intended word. For example, a person with dementia might say ``oven'' instead of ``sink'' or ``water'' instead of ``cookies.''\\
Discourse Impairment: There may be a noticeable decline in the ability to organize narrative speech. This can include tangential speech, difficulty maintaining a topic, and problems with coherence and cohesion. The person might jump from one idea to another without clear connections, or they might provide too much or too little information about the picture.\\\\
For each of these indicators, and for the following transcript of a Cookie Theft Picture description task, please indicate how much each indicator is fulfilled by the transcript. Give numbers between 1 (not at all fulfilled) and 7 (very strongly fulfilled). In addition, give 1-3 examples from the text for your assessment (in brackets, mark verbatim quotations with "). Give only the indicator name,  the assessment (number between 1 and 7), examples from the text (in brackets), nothing else, and no explanation.\\\\
\{transcript\}} \\\\
\textbf{GPT-4 response:} \texttt{Word-Finding Difficulties (Anomia): 5 (``no I can't no I can't get this very well, clear'', ``it's mm well somebody's drying dishes'')} \\
\texttt{Syntactic Simplification: 4 (use of simple sentences and phrases like ``I see a tad bit'', ``someone's standing on a stool'')} \\
\texttt{Discourse Impairment: 5 (``no I don't see anything else going on over here'', disjointed narrative with repetitions and lack of coherence)} \\
\texttt{Impoverished Vocabulary: 4 (repeated use of ``somebody'', ``something'', ``drying dishes'')} \\
\texttt{Semantic Paraphasias: 2 (no clear evidence of semantic paraphasias, but some difficulty in expression)}

\end{minipage}
    }
    \caption{GPT \emph{Prompt 2}, used to extract feature values for our GPT features, and the GPT response for an AD patient in our dataset. The verbatim transcript is replaced by a placeholder {\footnotesize\texttt{\{transcript\}}}. The system message is not shown here, but provided in Appendix~\ref{appendix:gpt-prompts}.}
    \label{fig:gpt-prompt}
\end{figure}

\subsection{Validation of GPT features}

\subsubsection{Clinical validation} 
For each feature, we perform a group comparison between AD and control, calculating Cohen's \emph{d} (effect size, \citealp{cohen1988}). In addition, we statistically test whether values in AD are significantly larger than values in the Control group using a Mann-Whitney U Test \cite{mann1947test}, as it is a non-parametric test suitable for ordinal features.

\subsubsection{Validation of \emph{Word-Finding Difficulties}}

While GPT readily quantifies the five studied indicators, it is unclear whether these assessments indeed capture the speech characteristics in question. For the feature \emph{Word-Finding Difficulties (Anomia)}, we try to validate this by comparing it to a deterministic proxy measure and to human ratings.

\paragraph{Validation against proxy measure:}  
We compute a proxy feature \emph{disfluency ratio}, defined as the number of disfluencies divided by the total number of spoken words. To count the number of disfluencies, we use all relevant special transcription markers provided by the CHAT format: fragmented words (e.g. ``coo'' instead of ``cookie''), filler words (e.g. ``uhm''), as well as explicitly coded repetitions (\texttt{[/]}), revisions (\texttt{[//]}), and pauses (\texttt{(.)}, \texttt{(..), \texttt{(...)}}). This proxy feature is our best attempt to quantify word-finding difficulties using the available annotations and traditional methods of computational linguistics. For this reason, we expect the GPT feature \emph{Word-Finding Difficulties (Anomia)} to be more highly correlated to \emph{disfluency ratio} than to other features, indicating that the two features are related, and increasing our confidence in GPT successfully capturing the speech characteristic in question.

\paragraph{Agreement to human raters:} We have asked two specialists (a psychologist and a speech therapist) to rate the word-finding difficulties for all subjects in our dataset, given both the audio recording and the manual transcript. These human ratings serve as a gold standard that we aim for our GPT-derived feature \emph{Word-Finding Difficulties (Anomia)} to approximate.  Given the subjectivity of the task, two human  raters will never agree perfectly. We quantify the amount of human (dis)agreement (i.e.\ the inter-rater reliability) using an intraclass correlation coefficient between these two raters (Case \emph{ICC(3,1)} according to \citet{shrout1979intraclass}) and corresponding 95\% confidence intervals\footnote{We use the \texttt{intraclass\_corr} method of Python's \texttt{pingouin} library v.\ 0.5.4 to calculate ICC and corresponding confidence intervals.}.
In addition, we calculate the ICC between the GPT feature values (on manual transcripts) and the human ratings, assessing how strongly GPT agrees with the human raters. The human ICC serves as an upper limit of agreement that can be expected between GPT (or any automatic measure) and human ratings.

\begin{figure*}[ht]
  \begin{minipage}[c]{.25\linewidth}
    \includegraphics[width=1.1\linewidth]{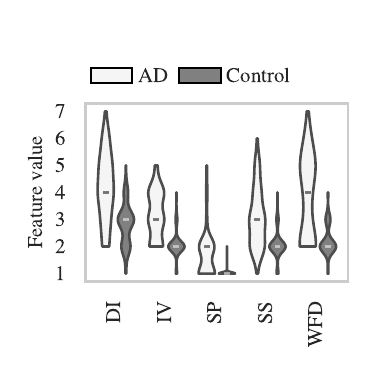}
    \vspace{-0.75cm}
  \end{minipage}
  \begin{minipage}[c]{.75\linewidth}
    \centering
    \scalebox{0.85}{
\begin{tabular}{l|cc|c|c}
\toprule
& \shortstack{\textbf{Control}\\\emph{(n=78)}} & \shortstack{\textbf{AD}\\\emph{(n=78)}} & \shortstack{\textbf{Cohen's}\\ \textbf{\emph{d}}} & \shortstack{\textbf{\emph{p}-}\\ \textbf{value}} \\
\midrule
Discourse Impairment (DI)              & 2.8 $\pm$ 0.8 & 4.1 $\pm$ 1.3 & 1.25 & 6.3e-11 \\
Impoverished Vocabulary (IV)           & 2.1 $\pm$ 0.5 & 3.2 $\pm$ 0.9 & 1.55 & 1.9e-15 \\
Semantic Paraphasias (SP)              & 1.0 $\pm$ 0.1 & 1.7 $\pm$ 0.8 & 1.12 & 7.8e-13 \\
Syntactic Simplification (SS)           & 2.0 $\pm$ 0.5 & 3.1 $\pm$ 1.1 & 1.26 & 1.9e-12 \\
Word-Finding Difficulties (WFD) & 2.2 $\pm$ 0.6 & 3.8 $\pm$ 1.5 & 1.34 & 8.8e-11 \\
\bottomrule
\end{tabular}
}
    \vspace{-0.75cm}
  \end{minipage}
    \caption{Clinical validation results for \emph{GPT features}. \textbf{Left:} Violin plots depicting the distribution of GPT feature values. Inner lines indicate median values. \textbf{Right:} Mean and standard deviation of the feature values for AD and control groups. We report Cohen's \emph{d} as a metric of effect size, as well as \emph{p}-values of the Mann-Whitney U Test.}
    \label{fig:clincial_validation}
\end{figure*}

\begin{table*}[!t]
    \centering
    \scalebox{0.9}{
\begin{tabular}{lccc}
    \toprule

                          & \textbf{Manual transcripts}                                 & \textbf{Google Speech ASR}                                 & \textbf{Whisper ASR}                                       \\ 
                          \midrule
Fine-tuned GPT          & $0.886$ ${\scriptstyle [0.831, 0.936] }$           & $0.862$ ${\scriptstyle [0.792, 0.918] }$          & $0.831$ ${\scriptstyle [0.760, 0.898] }$          \\
GPT features + RF         & $0.767$ ${\scriptstyle [0.683, 0.838] }$           & $0.760$ ${\scriptstyle [0.680, 0.835] }$          & $0.702$ ${\scriptstyle [0.615, 0.780] }$          \\
Established features + RF & $0.885$ ${\scriptstyle [0.829, 0.934] }$           & $0.893$ ${\scriptstyle [0.840, 0.939] }$          & $0.874$ ${\scriptstyle [0.811, 0.925] }$          \\
Established+GPT + RF      & $\textbf{*0.931}$ ${\scriptstyle [0.890, 0.962] }$ & $\textbf{0.900}$ ${\scriptstyle [0.857, 0.941] }$ & $\textbf{0.886}$ ${\scriptstyle [0.829, 0.934] }$ \\
\bottomrule
\end{tabular}
    }
    \caption{AUROC results, on 10-fold cross validation (CV). Results are given for manual transcripts, as well as Google Speech and Whisper ASR transcripts. Result estimates are provided with bootstrap confidence intervals. Bolt numbers mark the best result in each column. Asterisks in the last line indicate that results are significantly better than the line above, i.e.\  \emph{Established+GPT + RF} $>$ \emph{Established features + RF}. 
    }
    \label{tab:results}
\end{table*}

\subsection{AD classification and evaluation}
\noindent\textbf{Feature engineering + Random Forest:}
We train a Random Forest (RF) classifier\footnote{Using Python's \texttt{scikit-learn} library v1.2.2, with 500 estimators.} for binary classification (AD vs.\ control) on three sets of features: Our baseline \emph{Established features}, our \emph{GPT features}, and their union \emph{Established+GPT}. \\

\noindent\textbf{Fine-tuning GPT:} 
In addition to using GPT as a feature extractor in combination with RF, we directly fine-tune GPT to distinguish AD from control\footnote{We use GPT-4o for fine-tuning, as GPT-4 is not available for custom fine-tuning at the time of writing.}. This is conceptually similar to fine-tuning BERT, a common and successful approach to the AD detection task in prior work. For fine-tuning, we use a prompt asking GPT to decide whether a transcript comes from a person with AD or a healthy person, and ground truth completions ‘Dementia’ or ‘Healthy’. On test samples, the first token's log probability (which is provided by the fine-tuned model) is converted into a predicted AD probability, allowing the calculation of our metrics. Fine-tuning is orchestrated programmatically using the OpenAI API, with default hyperparameters\footnote{For an introduction to GPT finetuning, we refer the reader to the \href{https://platform.openai.com/docs/guides/fine-tuning}{OpenAI docs}. The Python code for our implementation can be accessed on our GitHub repository: \href{https://github.com/jheitz/coling2025_gpt_paper/blob/main/src/model/gpt_finetuning.py}{\texttt{src/model/gpt\_finetuning.py}}}. \\ 

\noindent\textbf{Evaluation:}  Classification performance is assessed using stratified 10-fold cross validation (CV) on the entire dataset (156 samples), with fixed random splits. We combine the test predictions of the 10 folds and report the area under the ROC curve (AUROC) on their union. To quantify uncertainty of results, we provide two-sided bootstrap confidence intervals (CI)\footnote{In this work, we estimate bootstrap confidence intervals (CI) using \texttt{stats.bootstrap} from Python's \texttt{scipy} library v1.11.3, with $1000$ bootstrap samples drawn from the dataset, and a confidence level of 0.95.}.

To statistically test if our \emph{GPT features} improve classification, we estimate a bootstrap CI of the performance difference $\delta_{{\scriptscriptstyle\text{AUROC}}}$  between \emph{Established features} and \emph{Established+GPT}, with  
\begin{equation}
\begin{split}
    \delta_{{\scriptscriptstyle\text{AUROC}}} = &\; \text{AUROC}_\text{Established+GPT} \\ 
    &- \text{AUROC}_\text{Established features}
\end{split}
\end{equation}
\emph{Established+GPT} is significantly better than \emph{Established features} if the entire CI of $\delta_{{\scriptscriptstyle\text{AUROC}}}$ is larger than $0$. 

The usefulness of individual features for classification is quantified using mean absolute SHAP values as a metric of feature importance \cite{lundberg2020shap}.

\subsection{Sensitivity analysis}
\label{sec:sensitivity-analysis}
\paragraph{Prompt and random seed:} LLMs are known to be sensitive to slight changes in prompts \cite{errica2024sensitivity, gan2023sensitivity, atil2024llmstability}. To analyse how stable our GPT feature values are subject to this \emph{prompt sensitivity}, we created two variations of \emph{Prompt 2}, where instructions are given in other words while keeping their meaning (full prompts are given in Appendix~\ref{appendix:alternative-prompts}). We use these to extract two additional sets of GPT feature values. We compute intraclass correlation coefficients (Case \emph{ICC(2,1)} according to \citet{shrout1979intraclass}) to assess how strongly result of different prompt versions agree. In addition, we calculate the difference of feature value when using the modified prompt compared to the original \emph{Prompt 2} for each feature and participant, and report their mean absolute difference (MD), defined as
\begin{equation}
 MD = \frac{1}{2 n} \sum_{v \in \{1,2\}} \sum_{i \in \{1, \dots, n\}} | \tilde{f_i^v} - f_i |
\end{equation}
where for participant $i$ among $n$ participants, $f_i$ refers to the original feature value and $\tilde{f_i^v}$ represents the feature value when using the alternative prompt $v$.

Analogously, we test the sensitivity of our GPT feature values to different random seeds in the OpenAI API call. \\

\paragraph{Number of GPT features:} We also perform a control analysis in which GPT was prompted to extract ten indicators instead of five, allowing us to evaluate the robustness of the feature selection process.\\

\section{Results}

\noindent\textbf{Comparison of ASR models:} Word Error Rates (WER) for both ASR models are similar, with the following median WER on the entire dataset (lower is better): Whisper: $0.35$ (AD: $0.43$, control: $0.31$), Google Speech: $0.37$ (AD: $0.40$, control: $0.30$). For AD classification (cf.\ results in Table~\ref{tab:results}), we observe a trend of Google Speech providing more useful transcripts than Whisper. \\

\noindent\textbf{Clinical validation of GPT features:} Validation results of our five \emph{GPT features} are presented in Figure~\ref{fig:clincial_validation}. We observe that GPT feature values are clearly higher for AD than control, with highly significant group differences (\emph{p}-values $< 10^{-10}$) and large effect sizes (Cohen's \emph{d} $> 1.1$). This confirms that all \emph{GPT features} capture language characteristics that are clinically relevant to distinguish individuals with AD from healthy controls. \\

\noindent\textbf{Validation of \emph{Word-Finding Difficulties}:}
The correlation between the proxy feature \emph{disfluency ratio} and \emph{Word-Finding Difficulties (Anomia)} is $0.63$, which is higher than the absolute correlation to any other linguistic feature ($\leq 0.55$, c.f.\ Appendix~\ref{appendix:feature-correlation} for a full correlation matrix). This strengthens our hypothesis that this GPT feature indeed captures difficulties in word-finding. \\
The intraclass correlation coefficient (ICC) between the two human raters quantifying word-finding difficulty is $0.55$ (CI: $0.43 - 0.65$), indicating moderate inter-rater reliability \cite{shrout1979intraclass}, which highlights the inherent subjectivity of assessing high-level concepts such as ``word-finding difficulty''. The ICC between GPT and human raters is $0.53$ (CI: $0.44 - 0.62$), with a confidence interval (CI) that overlaps with the human agreement. This indicates that GPT's assessment captures the speech characteristic in question as well as a human rater. It is important to note that the GPT assessment is based solely on the speech transcripts, whereas human raters had the advantage of basing their assessment on both audio and transcripts.  \\

\noindent\textbf{AD classification performance:} Classification results are given in Table~\ref{tab:results}.
We observe that our \emph{GPT features} alone perform worse than the \emph{Established features} (our baseline), when combined with RF. Fine-tuning GPT also does not outperform our baseline. However, RF with the combination of both established and GPT features (\emph{Established+GPT}) outperforms the established features. The differences are statistically significant on manual transcripts and stronger on manual than on ASR transcripts, but robust across all settings. Note that confidence intervals are relatively wide. This is a result of the small number of samples in our dataset, a main limitation of this work.  \\

\begin{table}[]
    \centering
    \scalebox{0.85}{
\begin{tabular}{lrr}
\toprule
\textbf{Feature name} & \multicolumn{2}{r}{\shortstack[r]{\textbf{Mean absolute}\\ \textbf{SHAP value}}} \\
\midrule
\multicolumn{2}{l}{Impoverished Vocabulary (\emph{GPT feature})} & 0.054 \\
\multicolumn{2}{l}{Word-Finding Difficulties (Anomia) (\emph{GPT})} & 0.039 \\
\multicolumn{2}{l}{Semantic Paraphasias (\emph{GPT feature})} & 0.038 \\
\multicolumn{2}{l}{avg\_word\_length (\emph{Established feature})} & 0.029 \\
\multicolumn{2}{l}{Syntactic Simplification (\emph{GPT feature})} & 0.028 \\
\multicolumn{2}{l}{Discourse Impairment (\emph{GPT feature})} & 0.027 \\
\multicolumn{2}{l}{adverb\_ratio (\emph{Established feature})} & 0.026 \\
\multicolumn{2}{l}{flesch\_kincaid (\emph{Established feature})} & 0.021 \\
\multicolumn{2}{l}{PRP\_ratio (\emph{Established feature})} & 0.019 \\
\multicolumn{2}{l}{n\_unique\_words (\emph{Established feature})} & 0.017 \\
\bottomrule
\end{tabular}
    }
    \caption{Feature importance of the top $10$ (out of $45$) features among \emph{Established+GPT}, based on results of 10-fold CV using manual transcripts. We provide mean absolute SHAP values.}
    \label{tab:feature_importance}
\end{table}

\noindent\textbf{Feature importance:} Table~\ref{tab:feature_importance} presents the $10$ most important features among \emph{Established+GPT}, according to mean absolute SHAP values \cite{lundberg2020shap}. We observe that GPT features are highly important, representing $5$ out of the top $6$ features.  \\

\noindent\textbf{Sensitivity analysis:} Table~\ref{tab:prompt_seed_sensitivity} displays the results for prompt and seed sensitivity. Low MD (approx.~$0.2$  on a scale from 1 -- 7) and high ICC (all~$>0.79$, considered excellent agreement \cite{cicchetti1994guidelines}) indicate that slight changes in prompts or random seeds have a low impact on the feature values. \\
Our control analysis extracting ten GPT features (instead of five) yielded a classification performance similar to our main results (results are listed in  Appendix Table~\ref{tab:results_control_anlaysis_10features}). \\

\begin{table}[!t]
    \centering
    \scalebox{0.85}{
\begin{tabular}{lcccc}
\toprule
\textbf{Sensitivity to different...} & \multicolumn{2}{c}{\textbf{Seeds}}  & \multicolumn{2}{c}{\textbf{Prompts}}  \\
 & MD & ICC & MD & ICC \\
\midrule
Discourse Impairment & 0.30 & 0.89 & 0.31 & 0.89 \\
Syntactic Simplification & 0.26 & 0.85 & 0.37 & 0.80 \\
Impoverished Vocabulary & 0.16 & 0.90 & 0.29 & 0.83 \\
Word-Finding Difficulties & 0.10 & 0.97 & 0.18 & 0.96 \\
Semantic Paraphasias & 0.09 & 0.79 & 0.10 & 0.81 \\
\midrule
\emph{Average} & 0.18 & 0.88 & 0.25 & 0.86 \\
\bottomrule
\end{tabular}
}
    \caption{Sensitivity of GPT feature values to changes in prompt wording and random seed: We report mean absolute difference (MD) and intraclass correlations (ICC) of feature values (on a scale from 1~--~7) when using alternative prompt wording / seed, as compared to the value using the original prompt / seed.}
    \label{tab:prompt_seed_sensitivity}
\end{table}

\noindent\textbf{Running time:} Running time is dominated by ASR, taking approx.\ \SI{55}{\minute} (Whisper) or \SI{30}{\minute} (Google Speech) for the entire dataset. Feature extraction and RF model training times sum up to less than \SI{10}{\minute}. GPT fine-tuning takes around \SI{15}{\minute} per split, with limited parallelization allowed by the API. Note that the inference time per individual is short enough to be deployed in a potential real-time application.  \\

\begin{table*}
    \centering
    \scalebox{0.68}{
\begin{tabular}{l|p{6cm}|p{15.5cm}}
&\textbf{Example transcript} & \textbf{GPT-4 response to \emph{Prompt 2}} \\
\midrule
\texttt{1} & \texttt{there's something has to be where the water goes down over} & \texttt{Word-Finding Difficulties (Anomia): 6 (Examples: ``something has to be where the water goes down over'' - struggles to find the word ``sink'' or ``tap'') [...]} \\
\midrule
\texttt{2} &\texttt{what do what do you call this ? the plate a plate ?} & \texttt{Word-Finding Difficulties (Anomia): 7 (Examples: ``what do what do you call this? the plate a plate?'') [...]} \\
\midrule
\texttt{3} &\texttt{the mother's washing dishes and water's spilling over} & \texttt{Word-Finding Difficulties (Anomia): 1 (No evidence of word-finding difficulties; the speaker uses specific terms like ``mother'' and ``washing dishes.'') [...]} \\
\bottomrule
\end{tabular}
    }
    \caption{GPT feature extraction responses for the feature \emph{Word-Finding Difficulties (Anomia)} on excerpts from transcripts in the dataset as a toy examples. Two examples represent high, one exemplifies low word-finding difficulty.}
    \label{tab:gpt-response-examples}
\end{table*}

\section{Discussion and Conclusion}

In this study, we have harnessed the capabilities of GPT-4 to extract five semantic features from transcripts of spontaneous speech, which were then utilized to complement existing linguistic features within an Alzheimer's Disease (AD) detection framework. The extracted features quantify known symptoms of AD in speech, but previous efforts in linguistic feature engineering have failed to capture them due to their complex and elusive nature.

For example, the feature \emph{Word-Finding Difficulties (Anomia)} is associated with AD \cite{rohrer2008word}, but its calculation from a given transcript is non-trivial and we are not aware of any existing linguistic feature attempting to do so. We have attempted to construct a rule-based deterministic proxy of this characteristic (\emph{disfluency ratio}, based on manual annotations of disfluencies) and showed that it is correlated highly with this GPT feature. In addition, we have demonstrated an agreement of this GPT feature's values with human raters. Compared to existing linguistic features, we believe that the GPT-extracted \emph{Word-Finding Difficulties (Anomia)} captures a richer concept, and a group comparison between AD and control as well as our SHAP feature importance analysis confirm the utility of this feature in AD detection. Table~\ref{tab:gpt-response-examples} displays three example transcripts, and the corresponding GPT-4 output for this feature. The difference in word-finding difficulty between these examples is apparent, but goes beyond the expressiveness of traditional linguistic features. Similar arguments can be made for all five features included in this study.

The GPT-generated features represent high-level speech alternations which are easier to grasp intuitively than many existing highly specific linguistic features. GPT-4 not only quantifies these features on a scale from 1 to 7, but also complements its assessment with explanatory notes or selected transcript excerpts (see examples responses in Figure~\ref{fig:gpt-prompt} and Table~\ref{tab:gpt-response-examples}). A potential application could provide these details to clinicians and patients. The transparent derivations and the intuitive interpretation of the features are a step towards better explainability, which is a crucial property of AI in medical applications, as it augments transparency, promotes trust of both clinicians and patients, and satisfies legal requirements such as the EU's ``AI Act''. We want to stress that explainability of this sort is a significant advantage of our feature-based approach compared to a fine-tuned language model such as GPT (as presented in this paper) or BERT (as frequently done in prior work). While fine-tuning is elegant and can be high-performing, explainability remains difficult, hindering a potential application in clinical practice.

Our results demonstrate how the GPT-generated features alone as well as the fine-tuning of GPT produce sub-optimal classification performance. This is in line with prior research employing GPT for AD classification, where results were unsatisfactory \cite{balamurali2024performance, wang2023can}. We hypothesize that fine-tuning does not work better because many low-level features of language  (e.g.\ based on word and letter counts) that are useful to distinguish AD from control cannot be effectively extracted by an LLM. For example, it has been shown that LLMs are unable to count letters \cite{zhang2024large}, making it impossible to extract a feature such as \emph{Average Word Length}, which is among our most important features (cf.\ Table~\ref{tab:feature_importance}). 
However, the combination of established linguistic with GPT-generated features produces a system that significantly outperforms prior feature-based approaches. 
This demonstrates that the complex semantic patterns identified by GPT encompass additional significant information beyond the reach of simpler, established features, which fail to capture such depth.
Moreover, they add a significantly different perspective: The maximal absolute correlation between a GPT feature and any other included linguistic feature is $0.55$ and thus rather low (a full correlation matrix is given in Appendix~\ref{appendix:feature-correlation}). Furthermore, the new GPT features are clinically relevant (cf.\ Figure~\ref{fig:clincial_validation}) and show high feature importance compared to established features (cf.\ Table~\ref{tab:feature_importance}), with $5$ out of the top $6$ most important features given by GPT. This further strengthens the observation that they indeed capture meaningful symptoms of AD.

Our approach is robust to slight variations in prompts or random seeds and is effective with both manual transcripts of spontaneous speech and ASR transcripts, where we recommend the use of Google Speech, as it results in better AD classification performance than Whisper. We suspect the reason to be a stronger use of a language model in Whisper's decoding pipeline compared to Google Speech, smoothing ASR outputs in a way that removes details (e.g.\ repetitions) from the transcripts that prove valuable to distinguish AD from control. The strong performance on ASR transcripts could enable the use of our approach in a fully automatic, low-cost, and real-time system. This makes it suitable for large-scale epidemiological studies, or might allow the identification of AD subgroups based on linguistic characteristics (similar to \citealp{park2017robust}), paving the way for more tailored cognitive training interventions.

In addition to the presented method of leveraging GPT as a feature extractor and fine-tuning GPT for direct AD classification, we have experimented with other approaches of harnessing GPT for AD classification based on spontaneous speech, including direct zero-shot prediction of AD vs.\ control (instead of feature extraction), and the use of GPT-4o instead of GPT-4. These additional experiments showed no improvements over the presented methods (results are given in Appendix~\ref{appendix:additional-results}). In this work, we show no detailed results for a fine-tuned BERT model, but results reported in our previous work are inferior to the presented method here \cite{heitz2024influence}. \\

\noindent\textbf{Conclusion:} In this study, we employed GPT-4 to extract linguistic features capturing known alterations in AD speech from both manually transcribed and automatically generated transcripts of spontaneous speech, integrating these within an end-to-end AD detection framework. We demonstrated that the inclusion of complex GPT-derived features enhances performance beyond what is achievable with traditional linguistic features alone, surpassing a fine-tuned GPT model. Furthermore, these new features are accompanied by explanatory snippets extracted from the transcripts, contributing to the advancement of more interpretable AI within a medical setting. Our innovative approach of combining GPT-generated features with an established classification pipeline is a novel and effective application of recent advancements in LLMs for AD speech analysis.

\section{Limitations}

The primary limitation of this study is the size and diversity of the dataset. Although we are convinced that the usefulness of our new GPT features would generalize to larger and more diverse datasets, further research is required to strengthen our findings and effectively test for potential biases. 

Moreover, our approach relies on GPT, a technology controlled by a commercial organization. Its large-scale deployment could entail economic and ethical risks, and may be affected unpredictably by future developments. However, powerful open-source alternatives to GPT are available and could be explored instead.

Our approach is based on transcripts of speech, ignoring additional information contained in the audio signal. While prior research has shown that these linguistic features are more useful than audio features \cite{cummins2020comparison}, clinicians assessing speech disorders rely on both modalities, capturing the entirety of patients' speech. Future work should thus focus on multi-modal approaches -- recent developments on multi-modal LLMs are promising foundational steps into this direction.

More broadly, we recognize substantial potential in utilizing LLMs, such as GPT, to explain the outputs of existing predictive models. This approach can enhance the acceptance of machine learning models within the medical field by bridging the communication gap between complex (difficult to understand) models and clinicians without technical expertise.

\bibliography{custom}

\appendix
\section{Established linguistic features}
\label{appendix:features}

\begin{table*}
    \centering
    \scalebox{0.63}{
    \begin{tabular}{r|p{5.5cm}|p{8.7cm}|p{6.6cm}}
    Group & Feature Name & Description & Used by prior research \\
    \toprule
        \texttt{SYN/P} & \texttt{pronoun\_noun\_ratio} & Ratio of pronouns to nouns & \citet{fraser2016linguistic, balagopalan2020bert, liu2021automatic} \\    \hline
        \texttt{SYN/P} & \texttt{verb\_noun\_ratio} & Ratio of verbs to nouns & \citet{liu2021automatic} \\    \hline
        \texttt{SYN/P} & \texttt{subordinate\_coordinate\_ conjunction\_ratio} & Ratio of subordinate to coordinate conjunctions  & \citet{parsapoor2023performance} \\    \hline
        
        \texttt{SYN/P} & \texttt{adverb\_ratio} & Ratio of adverbs to all words & \citet{fraser2016linguistic, balagopalan2020bert, tang2023explainable}\\    \hline
        \texttt{SYN/P} & \texttt{noun\_ratio} & Ratio of nouns to all words & \citet{fraser2016linguistic, diaz2022increasing, tang2023explainable}\\    \hline
        \texttt{SYN/P} & \texttt{verb\_ratio} & Ratio of verbs to all words & \citet{fraser2016linguistic, tang2023explainable} \\    \hline
        \texttt{SYN/P} & \texttt{pronoun\_ratio} & Ratio of pronouns to all words & \citet{balagopalan2020bert, tang2023explainable}\\    \hline
        \texttt{SYN/P} & \texttt{personal\_pronoun\_ratio} & Ratio of personal pronouns to all words & \citet{balagopalan2020bert} \\    \hline
        \texttt{SYN/P} & \texttt{determiner\_ratio} & Ratio of determiners to all words & \citet{diaz2022increasing}\\    \hline
        \texttt{SYN/P} & \texttt{preposition\_ratio} & Ratio of prepositions to all words & \citet{tang2023explainable} \\    \hline
        \texttt{SYN/P} & \texttt{verb\_present\_participle\_ratio} & Ratio of verb (present participle) to all words & \citet{balagopalan2020bert, diaz2022increasing}\\    \hline
        \texttt{SYN/P} & \texttt{verb\_modal\_ratio} & Ratio of modal verbs to all words  &  \citet{diaz2022increasing}\\    \hline
        \texttt{SYN/P} & \texttt{verb\_third\_person\_ singular\_ratio} & Ratio of verbs in 3rd person singular to all words & \citet{fraser2016linguistic} \\    \hline
        
        \texttt{SYN/P} & \texttt{propositional\_density} & Based on POS tags, according to \citet{parsapoor2023performance} & \citet{parsapoor2023performance, eyigoz2020linguistic}\\    \hline
        \texttt{SYN/P} & \texttt{content\_density} & Based on POS tags, according to \citet{parsapoor2023performance} & \citet{parsapoor2023performance, diaz2022increasing, tang2023explainable} \\   
    \midrule
        \texttt{SYN/C} & \texttt{NP$\rightarrow$PRP} & \multirow{12}{8.7cm}{Count of respective context-free grammar (CFG) production rules according to constituency parsing} & \citet{fraser2016linguistic}\\     \cline{1-2}\cline{4-4} 
         \texttt{SYN/C} & \texttt{ADVP$\rightarrow$RB} & & \citet{fraser2016linguistic, balagopalan2020bert} \\     \cline{1-2}\cline{4-4}
         \texttt{SYN/C} & \texttt{NP$\rightarrow$DT\_NN} &  & \citet{fraser2016linguistic} \\     \cline{1-2}\cline{4-4}
         \texttt{SYN/C} & \texttt{ROOT$\rightarrow$FRAG} &  & \citet{fraser2016linguistic} \\     \cline{1-2}\cline{4-4}
         \texttt{SYN/C} & \texttt{VP$\rightarrow$AUX\_VP} &  & \citet{fraser2016linguistic}\\     \cline{1-2}\cline{4-4}
         \texttt{SYN/C} & \texttt{VP$\rightarrow$VBG} &  & \citet{fraser2016linguistic} \\     \cline{1-2}\cline{4-4}
         \texttt{SYN/C} & \texttt{VP$\rightarrow$VBG\_PP} &  & \citet{fraser2016linguistic}\\     \cline{1-2}\cline{4-4}
         \texttt{SYN/C} & \texttt{VP$\rightarrow$IN\_S} &  & \citet{fraser2016linguistic}\\     \cline{1-2}\cline{4-4}
         \texttt{SYN/C} & \texttt{VP$\rightarrow$AUX\_ADJP} &  & \citet{fraser2016linguistic}\\     \cline{1-2}\cline{4-4}
         \texttt{SYN/C} & \texttt{VP$\rightarrow$AUX} &  & \citet{fraser2016linguistic} \\     \cline{1-2}\cline{4-4}
         \texttt{SYN/C} & \texttt{VP$\rightarrow$VBD\_NP} &  & \citet{fraser2016linguistic} \\     \cline{1-2}\cline{4-4}
         \texttt{SYN/C} & \texttt{INTJ$\rightarrow$UH}&  & \citet{fraser2016linguistic} \\    \hline
        \texttt{SYN/C} & \texttt{NP\_ratio} & Ratio of noun phrases to all constituents & \citet{tang2023explainable}\\    \hline
        \texttt{SYN/C} & \texttt{PRP\_ratio} & Ratio of personal pronoun constituents to all constituents & \citet{tang2023explainable} \\    \hline
        \texttt{SYN/C} & \texttt{PP\_ratio} & Ratio of prepositional phrases to all constituents & \citet{fraser2016linguistic} \\    \hline
        \texttt{SYN/C} & \texttt{VP\_ratio} & Ratio of verb phrases to all constituents & \citet{fraser2016linguistic} \\     \hline
        \texttt{SYN/C} & \texttt{avg\_n\_words\_in\_NP} & Average number of words in noun phrase  & \citet{tang2023explainable} \\   
    \midrule
      \texttt{LEX} & \texttt{flesch\_kincaid} & The Flesch–Kincaid Grade Level Formula \citet{kincaid1975fleschkincaid}, a metric of readability. &  \citet{parsapoor2023performance} \\    \hline
     \texttt{LEX} & \texttt{avg\_word\_length} & Average letters per word  & \citet{fraser2016linguistic, balagopalan2020bert} \\    \hline
     \texttt{LEX} & \texttt{n\_words} & Nubmer of words in transcript & \citet{tang2023explainable, liu2021automatic, priyadarshinee2023alzheimer, diaz2022increasing} \\    \hline
     \texttt{LEX} & \texttt{n\_unique\_words} & Nubmer of unique words in transcript & \citet{priyadarshinee2023alzheimer, diaz2022increasing} \\     \hline
     \texttt{LEX} & \texttt{avg\_sentence\_length} & Average number of words per sentence & \citet{liu2021automatic}\\    \hline
     \texttt{LEX} & \texttt{words\_not\_in\_dict\_ratio} & Ratio of words not in English dictionary & \citet{fraser2016linguistic, balagopalan2020bert} \\    \hline
     \texttt{LEX} & \texttt{brunets\_index} & Brunét's index \citet{brunet1978brunetsindex}, a metric of lexical richness defined as $N^{V^{-0.165}}$, with $N$ the number of words and $V$ is the number of unique words & \citet{parsapoor2023performance, diaz2022increasing}\\    \hline
     \texttt{LEX} & \texttt{honores\_statistic} & Honoré Statistic \citet{honore1979honorestatistic}, a metric of lexical richness defined as $\frac{100 \log(N)}{1-V_1/V}$, with $N$ the number of words and $V$ is the number of unique words, and $V_1$ the number of unique words appearing once & \citet{fraser2016linguistic, tang2023explainable, parsapoor2023performance, diaz2022increasing}\\     \hline
     \texttt{LEX} & \texttt{ttr} & The type-token-ratio (TTR) \citet{miller1981ttr}, a measure of lexical diversity, defined as number of words devided by number of unique words. & \citet{liu2021automatic, diaz2022increasing} \\    \hline
     \texttt{LEX} & \texttt{mattr} & The moving-average type-token-ratio \citet{covington2010mattr} with window length 20. & \citet{diaz2022increasing} \\
    \midrule
    \texttt{REP} & \texttt{avg\_distance\_between\_ utterances} & Avg.\ cosine distance between utterances in transcript, a feature of repetitiveness, based on \citet{masrani2017detecting}'s implementation & \citet{fraser2016linguistic, balagopalan2020bert}\\    \hline
    \texttt{REP} &  \texttt{prop\_utterance\_dist\_below\_05} & Proportion of sentence pairs where cosine distance $\leq$ 0.5, based on \citet{masrani2017detecting}'s implementation & \citet{fraser2016linguistic, balagopalan2020bert}\\    \hline
    \bottomrule
    \end{tabular}
    }
    \caption{Table of all features within \emph{Established features}. Original source of this table: \citet{heitz2024influence}. Feature groups are: Syntactic features based on POS tags \texttt{SYN/P} or grammatical constituents \texttt{SYN/C}, lexical features \texttt{LEX}, and features of repetitiveness \texttt{REP}.}
    \label{tab:features}
\end{table*}

As a baseline set of linguistic features (which we call \emph{Established features} in this paper), we selected $40$ features motivated by their success in previous approaches on AD detection from spontaneous speech. Our selection of features is identical to our previous work \cite{heitz2024influence}. For convenience, we have reprinted the table of all features and their definitions in Table~\ref{tab:features}.

To arrive at our selection, we included all linguistic features used in \citet{fraser2016linguistic, balagopalan2020bert, parsapoor2023performance, liu2021automatic, syed2021automated, priyadarshinee2023alzheimer, eyigoz2020linguistic, diaz2022increasing, tang2023explainable}, if they were either a) present in at least two of these studies, or b) identified as important according to feature importance or a statistical test. Among these, we excluded all features where either a) the provided description was insufficient for reimplementation, or b) the feature values were all constant in our dataset, which is the case for some features based on grammatical constituents that were not found in our dataset.

We use the Stanza NLP library \cite{qi2020stanza} for constituency parsing and part-of-speech (POS) tagging\footnote{Version 1.5.0}. The code of our implementation can be accessed on our GitHub repository\footnote{\href{https://github.com/jheitz/coling2025_gpt_paper}{https://github.com/jheitz/coling2025\_gpt\_paper}}.

\section{Feature correlation}
\label{appendix:feature-correlation}

\begin{figure}[h!]
  \centering
  \includegraphics[width=\linewidth]{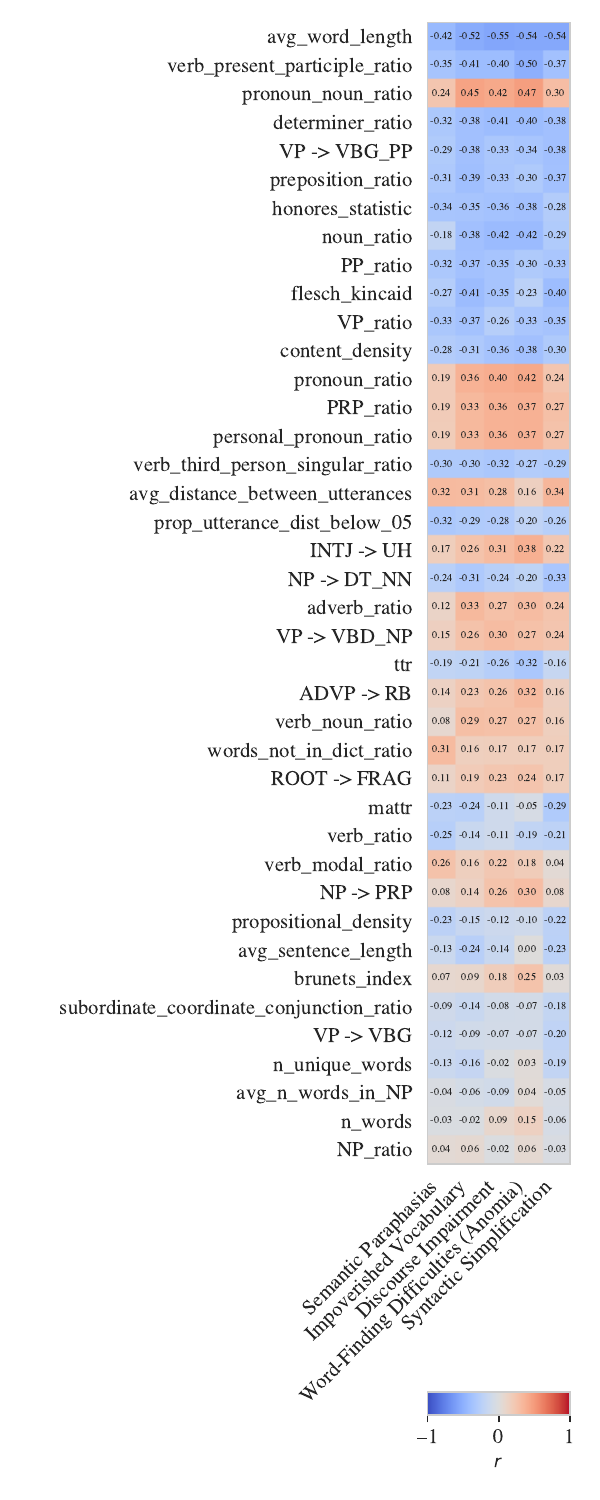}
  \caption{Feature correlation between \emph{GPT features} (on the x axis) and \emph{Established features} (on the y axis).}
  \label{fig:feature_correlation}
\end{figure}

Figure~\ref{fig:feature_correlation} displays the correlation between \emph{GPT features} and \emph{Established features}. The maximal absolute correlation is $0.55$ between \emph{Discourse Impairment} and \emph{Average Word Length}. This demonstrates that our new  \emph{GPT features} capture linguistic phenomena that are significantly different to existing linguistic features.

\section{GPT prompts}
\label{appendix:gpt-prompts}

In this section, we present the verbatim prompts introduced in Section~\ref{sec:gpt-prompts}: \emph{Prompt 1} has been used once, to initially find the five GPT features, without providing any transcripts. \emph{Prompt 2} is used for every transcript, to extract the value of these features.

\subsection{GPT Prompt 1 - Choice of features}
\label{sec:gpt-prompt-1}
GPT \emph{Prompt 1}, used to initially find relevant features, as well as the GPT-4 response, are given below:\\\\

\begin{scriptsize}
\noindent
\textbf{System message:} \texttt{You are a experienced doctor studying patients with Alzheimer's dementia. You know everything about this disease and how it affects language. You can analyze transcriptions of spontaneous speech and tell whether the speaker suffers from Alzheimer's Disease (dementia) or not. The transcriptions are typically created based on audio recordings of people describing the Cookie Theft Picture. In this task, participants are shown a drawing of a mother who is drying dishes next to the sink in the kitchen. She is not paying attention and has left the tap on.  As a  result, water  is overflowing from  the sink.  Meanwhile,  two children  are attempting to take cookies from a jar when their mother is not looking. One of the children, a boy, has climbed onto a stool to get up to the cupboard where the cookie jar is stored. The stool is rocking precariously. The other child, a girl, is standing next to the stool and has her hand outstretched ready to be given cookies.}\\\\
\textbf{Prompt 1:} \texttt{Name 5 good indicators that you would examine in these transcriptions of spontaneous speech, pointing towards Dementia} \\\\
\textbf{GPT-4 response:} \texttt{When analyzing transcriptions of spontaneous speech, particularly in the context of the Cookie Theft Picture description task, there are several linguistic indicators that may suggest the presence of Alzheimer’s dementia. Here are five key indicators to look for:}

\begin{itemize}
    \item \texttt{Word-Finding Difficulties (Anomia): Individuals with Alzheimer’s often have trouble finding the right words. This can manifest as frequent pauses, use of non-specific words like "thing” or "stuff,” or circumlocutions (talking around the word without being able to name it). For example, instead of saying "stool,” a person with dementia might say "the thing you sit on.”}
    \item \texttt{Impoverished Vocabulary: The range of words used by someone with Alzheimer’s may become limited. Their language may seem less rich and more repetitive, with a reliance on common and general terms rather than specific nouns or verbs.}
    \item \texttt{Syntactic Simplification: The complexity of sentence structures may decline. People with dementia might use simpler, shorter sentences and may make more grammatical errors. They might avoid complex grammatical constructions like subordination or relative clauses.}
    \item \texttt{Semantic Paraphasias: This refers to the use of incorrect words that are semantically related to the intended word. For example, a person with dementia might say "oven” instead of "sink” or "water” instead of "cookies.”}
    \item \texttt{Discourse Impairment: There may be a noticeable decline in the ability to organize narrative speech. This can include tangential speech, difficulty maintaining a topic, and problems with coherence and cohesion. The person might jump from one idea to another without clear connections, or they might provide too much or too little information about the picture.}
\end{itemize}

\noindent
\texttt{It’s important to note that while these indicators can suggest the presence of Alzheimer’s dementia, they are not definitive on their own. A thorough assessment by a healthcare professional, including a detailed medical history, cognitive testing, and possibly neuroimaging, is necessary for a formal diagnosis. Additionally, these symptoms can vary widely among individuals and may be influenced by other factors such as education level, language proficiency, and cultural background.}

\end{scriptsize}

\begin{table*}[!th]
    \centering
    \scalebox{0.9}{
\begin{tabular}{lccc}
    \toprule

                          & \textbf{Manual transcripts}                                 & \textbf{Google Speech ASR}                                 & \textbf{Whisper ASR}                                       \\ 
                          \midrule
Established+GPT-4o + RF    & $0.910$ ${\scriptstyle [0.859, 0.950] }$  & $
0.899$ ${\scriptstyle [0.850, 0.939] }$ & $0.872$ ${\scriptstyle [0.807, 0.922] }$          \\ 
GPT-4 zero shot & $0.827$ ${\scriptstyle [0.759, 0.888] }$ & $0.750$ ${\scriptstyle [0.674, 0.821] }$ & $0.740$ ${\scriptstyle [0.662, 0.811] }$ \\
GPT-4o zero shot & $0.677$ ${\scriptstyle [0.596, 0.760] }$ & $0.646$ ${\scriptstyle [0.563, 0.717] }$ & $0.663$ ${\scriptstyle [0.575, 0.746] }$ \\

\bottomrule
\end{tabular}
    }
    \caption{Additional results for AD classification based on \emph{GPT features} extracting using GPT-4o (instead of GPT-4), as well as GPT zero-shot classification performance using GPT-4 and GPT-4o. This is an extension of Table~\ref{tab:results}.}
    \label{tab:additional-results}
\end{table*}

\subsection{GPT Prompt 2: Feature extraction}
\label{sec:gpt-prompt-2}

GPT \emph{Prompt 2} is used to extract feature values for our GPT features. In the following, the transcript text is replaced by the  placeholder {\footnotesize\texttt{\{transcript\}}}. This prompt is identical to the one presented in Figure~\ref{fig:gpt-prompt}, additionally including the system message: \\\\

\begin{scriptsize}

\noindent
\textbf{System message:} \texttt{You are an experienced doctor studying patients with Alzheimer's dementia. You know everything about this disease and how it affects language. You can analyze transcriptions of spontaneous speech and tell whether the speaker suffers from Alzheimer's Disease (dementia) or not. The transcriptions you will see have been created based on audio recordings of people describing the Cookie Theft Picture. In this task, participants are shown a drawing of a mother who is drying dishes next to the sink in the kitchen. She is not paying attention and has left the tap on.  As a  result, water  is overflowing from  the sink.  Meanwhile,  two children  are attempting to take cookies from a jar when their mother is not looking. One of the children, a boy, has climbed onto a stool to get up to the cupboard where the cookie jar is stored. The stool is rocking precariously. The other child, a girl, is standing next to the stool and has her hand outstretched ready to be given cookies. Examinees are instructed "Tell me everything you  see going  on  in this  picture”.} \\\\
\textbf{Prompt 2:} \texttt{Here are key indicators of Alzheimer's dementia in spontaneous speech:\\\\
Word-Finding Difficulties (Anomia): Individuals with Alzheimer’s often have trouble finding the right words. This can manifest as frequent pauses, use of non-specific words like "thing” or "stuff”, or circumlocutions (talking around the word without being able to name it). For example, instead of saying "stool”, a person with dementia might say "the thing you sit on".\\
Impoverished Vocabulary: The range of words used by someone with Alzheimer’s may become limited. Their language may seem less rich and more repetitive, with a reliance on common and general terms rather than specific nouns or verbs.\\
Syntactic Simplification: The complexity of sentence structures may decline. People with dementia might use simpler, shorter sentences and may make more grammatical errors. They might avoid complex grammatical constructions like subordination or relative clauses.\\
Semantic Paraphasias: This refers to the use of incorrect words that are semantically related to the intended word. For example, a person with dementia might say "oven” instead of "sink” or "water” instead of "cookies.”\\
Discourse Impairment: There may be a noticeable decline in the ability to organize narrative speech. This can include tangential speech, difficulty maintaining a topic, and problems with coherence and cohesion. The person might jump from one idea to another without clear connections, or they might provide too much or too little information about the picture.\\\\
For each of these indicators, and for the following transcript of a Cookie Theft Picture description task, please indicate how much each indicator is fulfilled by the transcript. Give numbers between 1 (not at all fulfilled) and 7 (very strongly fulfilled). In addition, give 1-3 examples from the text for your assessment (in brackets, mark verbatim quotations with "). Give only the indicator name,  the assessment (number between 1 and 7), examples from the text (in brackets), nothing else, and no explanation.\\\\
\{transcript\}} 

\end{scriptsize}

\section{Alternative prompts to assess \emph{prompt sensitivity}}
\label{appendix:alternative-prompts}
In the following, the two alternative versions of \emph{Prompt 2} are given. They are used in our prompt sensitivity analysis described in Section~\ref{sec:sensitivity-analysis}. \\

\begin{scriptsize}

\noindent
\textbf{Prompt 2 (Alternative 1):} \texttt{The following are important indicators of Alzheimer's dementia in spontaneous speech:\\\\
Word-Finding Difficulties (Anomia): Individuals with Alzheimer’s often have trouble finding the right words. This can manifest as frequent pauses, use of non-specific words like "thing” or "stuff”, or circumlocutions (talking around the word without being able to name it). For example, instead of saying "stool”, a person with dementia might say "the thing you sit on".\\
Impoverished Vocabulary: The range of words used by someone with Alzheimer’s may become limited. Their language may seem less rich and more repetitive, with a reliance on common and general terms rather than specific nouns or verbs.\\
Syntactic Simplification: The complexity of sentence structures may decline. People with dementia might use simpler, shorter sentences and may make more grammatical errors. They might avoid complex grammatical constructions like subordination or relative clauses.\\
Semantic Paraphasias: This refers to the use of incorrect words that are semantically related to the intended word. For example, a person with dementia might say "oven” instead of "sink” or "water” instead of "cookies.”\\
Discourse Impairment: There may be a noticeable decline in the ability to organize narrative speech. This can include tangential speech, difficulty maintaining a topic, and problems with coherence and cohesion. The person might jump from one idea to another without clear connections, or they might provide too much or too little information about the picture.\\\\
For each listed indicator and the provided transcript of a Cookie Theft Picture description task, please rate how well the transcript meets each indicator on a scale from 1 (not at all) to 7 (extremely well). Provide 1-3 text examples to support your rating (in brackets, enclosing direct quotes in quotation marks "). Include only the indicator name, the rating, and examples from the text; no additional explanations. For each indicator, use the form: Indicator name: 1-7 (text examples with "quotes").\\\\
\{transcript\}} 

\end{scriptsize}

\vspace{0.5cm}

\begin{scriptsize}

\noindent
\textbf{Prompt 2 (Alternative 2):} \texttt{The following indicators help to detect individuals with Alzheimer's dementia using transcripts of spontaneous speech:\\\\
Word-Finding Difficulties (Anomia): Individuals with Alzheimer’s often have trouble finding the right words. This can manifest as frequent pauses, use of non-specific words like "thing” or "stuff”, or circumlocutions (talking around the word without being able to name it). For example, instead of saying "stool”, a person with dementia might say "the thing you sit on".\\
Impoverished Vocabulary: The range of words used by someone with Alzheimer’s may become limited. Their language may seem less rich and more repetitive, with a reliance on common and general terms rather than specific nouns or verbs.\\
Syntactic Simplification: The complexity of sentence structures may decline. People with dementia might use simpler, shorter sentences and may make more grammatical errors. They might avoid complex grammatical constructions like subordination or relative clauses.\\
Semantic Paraphasias: This refers to the use of incorrect words that are semantically related to the intended word. For example, a person with dementia might say "oven” instead of "sink” or "water” instead of "cookies.”\\
Discourse Impairment: There may be a noticeable decline in the ability to organize narrative speech. This can include tangential speech, difficulty maintaining a topic, and problems with coherence and cohesion. The person might jump from one idea to another without clear connections, or they might provide too much or too little information about the picture.\\\\
For each indicator, and the subsequent Cookie Theft Picture description task transcript, assign a value between 1 (not fulfilled at all) and 7 (extremely well fulfilled). Include 1-3 illustrative excerpts from the transcript (mark exact quotes with quotation marks "). Include only the indicator name, the rating, and examples from the text; no additional explanations. For each indicator, use the form: Indicator name: 1-7 (text examples with "quotes").\\\\
\{transcript\}} 

\end{scriptsize}

\section{Additional results}
\label{appendix:additional-results}

Table~\ref{tab:additional-results} provides additional results for feature extraction based on GPT-4o (instead of GPT-4) as well as zero-shot classification performance, where GPT is provided with transcripts and is asked to classify them into AD or control, without having any training data.

To assess whether extracting more than five GPT features would further improve our results, we conducted a control analysis, where we queried GPT for ten (instead of five) indicators of AD in speech, and extracted corresponding GPT features values for each participant in our dataset. Results are given in Table~\ref{tab:results_control_anlaysis_10features}.

\begin{table}[]
    \centering
    \scalebox{0.9}{
    \begin{tabular}{lll}
\toprule
& \textbf{Manual transcripts} \\
\midrule
GPT(10) features + RF & $0.783$ ${\scriptstyle[0.700, 0.851]}$ \\
GPT(5) features + RF & $0.767$ ${\scriptstyle [0.683, 0.838] }$ \\
Establ+GPT(10) + RF & $0.905$ ${\scriptstyle[0.851, 0.949]}$ \\
Establ+GPT(5) + RF & $0.931$ ${\scriptstyle [0.890, 0.962] }$ \\
\bottomrule
\end{tabular}
    }
    \caption{10-fold cross validation (CV) results of our control analysis extracting ten instead of five GPT features. GPT(5) refers to the five features described in this paper (results as presented in Table~\ref{tab:results}), GPT(10) represents the alternative set of ten features.}
    \label{tab:results_control_anlaysis_10features}
\end{table}

\end{document}